\documentclass[10pt,twocolumn,letterpaper]{article}

\usepackage{cls/cvpr}
\usepackage{times}
\usepackage{epsfig}
\usepackage{amsmath}
\usepackage{amssymb}
\usepackage{siunitx}
\usepackage{mathtools}
\usepackage[caption=false]{subfig}

\usepackage[breaklinks=true,bookmarks=false]{hyperref}

\cvprfinalcopy

\def\cvprPaperID{****} % *** Enter the CVPR Paper ID here

\begin{document}

\title{SynthCity: A large scale synthetic point cloud}

\author{David Griffiths$^{1*}$, Jan Boehm${^1}$\\
$^{1}$University College London, Gower Street, London, UK\\
{\tt\small \{david.griffiths.16, j.boehm\}@ucl.ac.uk}
}

\maketitle

\begin{abstract}
With deep learning becoming a more prominent approach for automatic classification of three-dimensional point cloud data, a key bottleneck is the amount of high quality training data, especially when compared to that available for two-dimensional images. One potential solution is the use of synthetic data for pre-training networks, however the ability for models to generalise from synthetic data to real world data has been poorly studied for point clouds. Despite this, a huge wealth of 3D virtual environments exist which, if proved effective can be exploited. We therefore argue that research in this domain would be of significant use. In this paper we present SynthCity an open dataset to help aid research. SynthCity is a 367.9M point synthetic full colour Mobile Laser Scanning point cloud. Every point is assigned a label from one of nine categories. We generate our point cloud in a typical Urban/Suburban environment using the Blensor plugin for Blender.  
\end{abstract}

\section{Introduction}

One of the fundamental requirements for supervised deep learning are large, accurately labelled datasets. For this reason, progress in two-dimensional (2D) image processing is often largely accredited to the wealth of very large, high quality datasets such as ImageNet \cite{DengJ2009} (classification), COCO \cite{LinEtAlT2014} (object detection) and Pascal VOC \cite{EveringhamEtAlM2010} (segmentation). It is now common practice to pre-train Convolutional Neural Networks (CNN) on large datasets before fine-tuning on smaller domain specific datasets. Despite the large success of deep learning for 2D image processing, it is evident that automatic understanding for three-dimensional (3D) point cloud data is not as mature. We argue one of the reasons for this is the lack of training data at the scale of that available for 2D data.

\begin{figure}[t!]
	\centering
	
	\subfloat[]{
		\label{subfig:pc_labels}
		\includegraphics[width=0.48\textwidth]{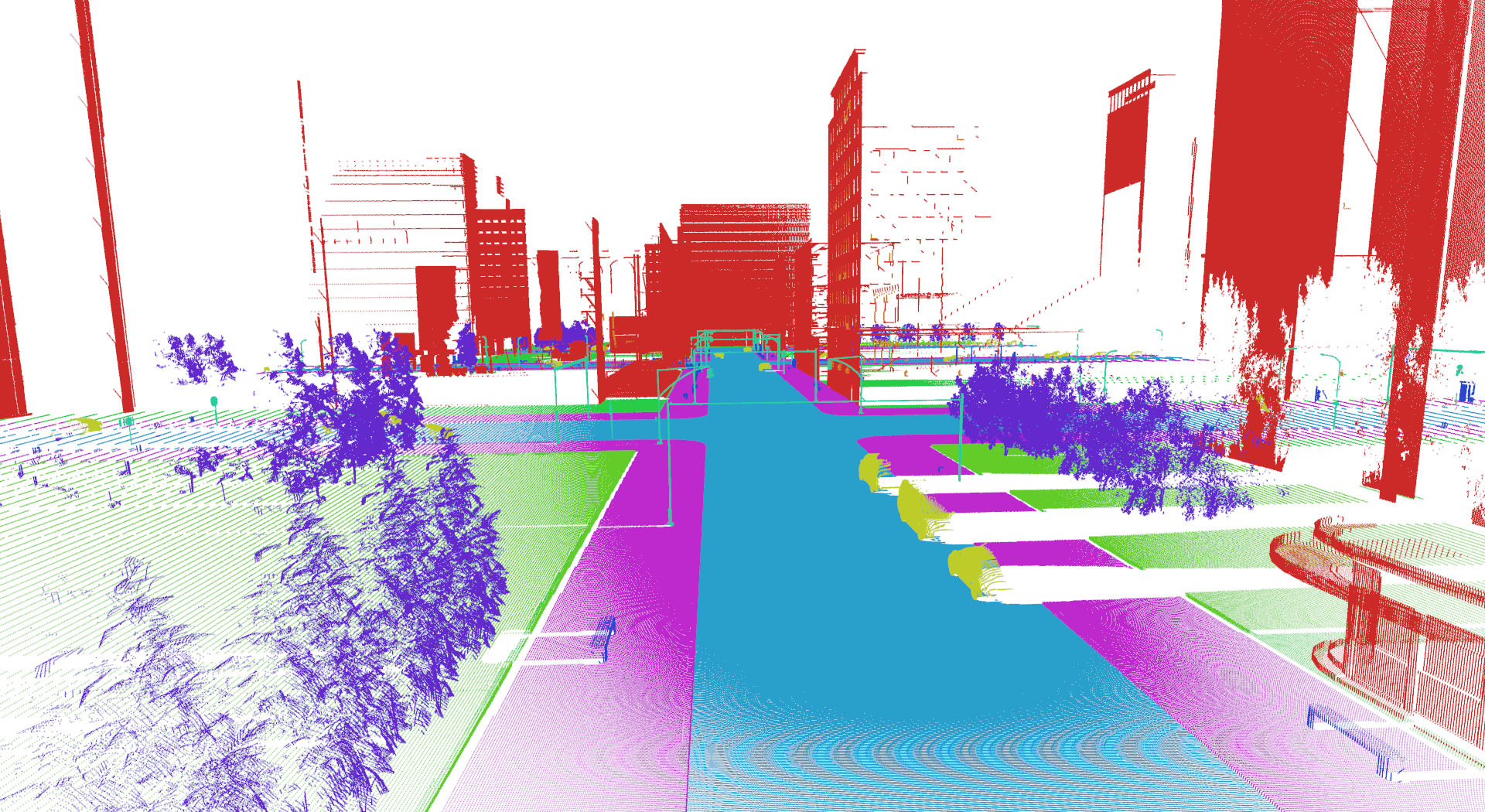}}
	
	\subfloat[]{
		\label{subfig:pc_rgb}
		\includegraphics[width=0.48\textwidth]{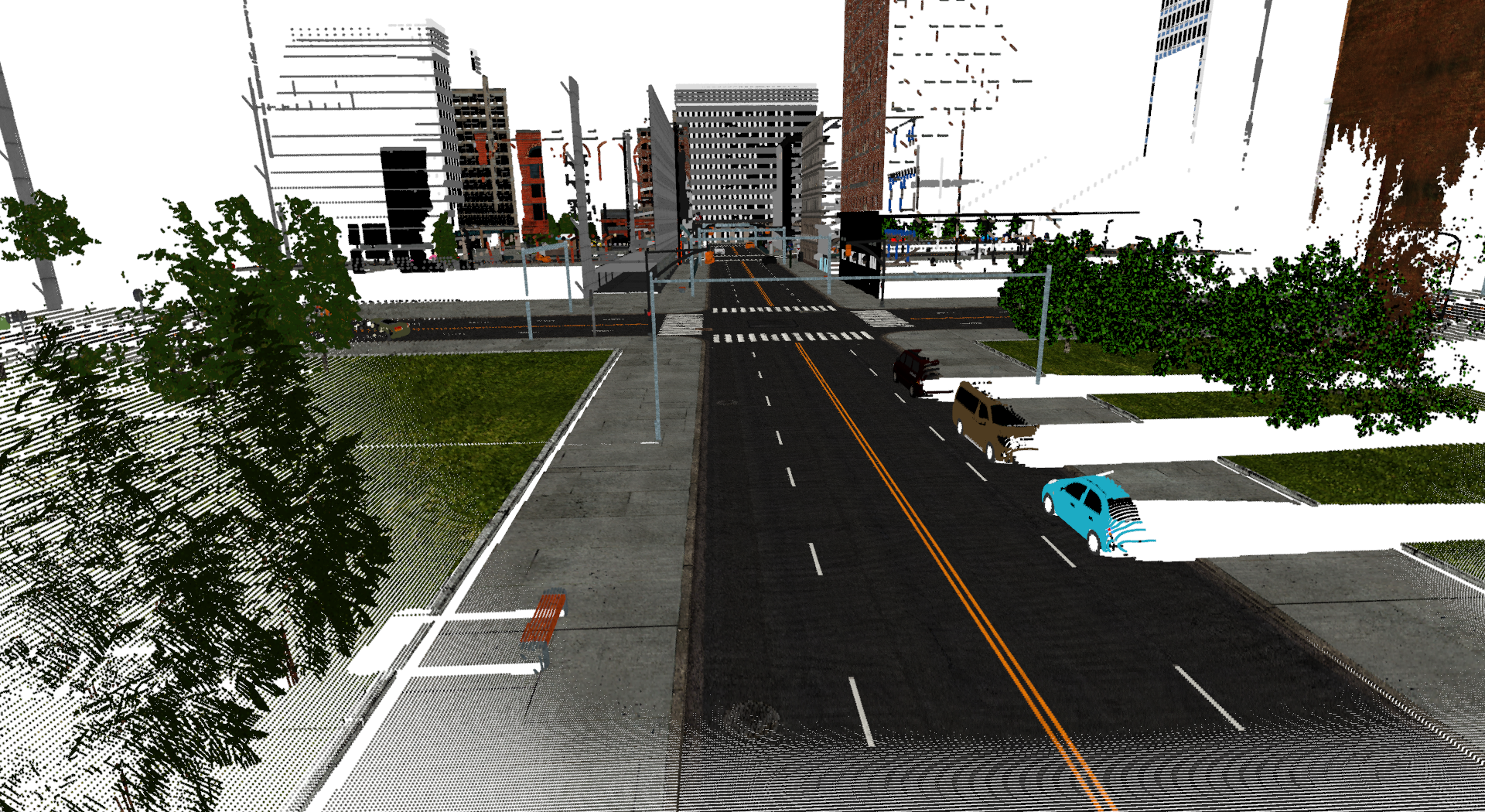}}
	
	\caption{Example of the SynthCity dataset displaying a) class labels and b) RGB values.}
	\label{fig:example_pc}
\end{figure}

A key reason for the lack of 3D training data is that naturally the amount of prepared labelled data decreases as the complexity of labelling increases. For example in 2D, single image classification (i.e. dog, car, cup etc.) is generally trivial and can therefore be carried out by large communities of untrained workers. Object detection requires more skill and has an added level of subjectivity. Segmentation again requires further precision, delicacy and involves more subjectivity. Per-point 3D segmentation requires highly skilled users and generating perfect labels for even the most advanced users is non-trivial. A potential solution to account for this is to synthetically generate training data (i.e. ShapeNet \cite{ChangEtAlA2015b}). Despite general success when pre-training 2D images on synthetic data and fine-tuning on real-world data, there has been very little research on this topic with respect to point cloud classification.

More so than 2D, 3D data benefits from a wealth of synthetic data in the form of virtual 3D environments generated for the purpose of gaming, virtual reality and scenario training simulators to name a few. However, the ability for deep learning networks to generalise from synthetic point clouds to real-world data is poorly studied, and as such the community risks missing out on a massive resource of data. To help address this we introduce \textit{SynthCity} an open, large scale synthetic point cloud of a typical urban/suburban environment. SynthCity is captured using a simulated Mobile Laser Scanner (MLS). MLS point cloud data capturing is being increasingly used due to its ability to easily cover large areas when compared to a Terrestrial Laser Scanner (TLS) and at a higher resolution than an Aerial Laser Scanner (ALS). However, whilst capturing large quantities of data is becoming more trivial, such large datasets are useless without the means to extract useful structured information from otherwise useless unstructured data. As such, progress in this field offers huge potential for a range of disciplines from city planning to autonomous driving.

The primary purpose of our dataset is therefore to offer an open dataset to aid further research assessing the potential of synthetic datasets for pre-training Deep Neural Networks (DNNs) for automatic point cloud labelling. We believe successful progression in this area could have potentially huge implications of the future of automatic point cloud labelling. Our dataset is available for download at \href{http://www.synthcity.xyz}{http://www.synthcity.xyz}.

\section{Related Work}

The need for outdoor labelled point clouds has been addressed by a range of researchers. Serna et al., \cite{SernaEtAlA2014} released the Paris-rue-Madame MLS dataset containing 20M points ($xy, y, z$ and reflectance), Vallet et al., \cite{ValletEtAlB2015} the iQmulus dataset containing 300M points ($x,y,z,$, time, reflectance and number of echoes) and Roynard et al., \cite{RoynardEtAlX2017} the Paris-Lille-3D containing 143.1M points ($x,y,z,$, scanner $x,y,z$, gps time, reflectance). However, many caveats exist within these datasets. For example, Paris-rue-Madame, whilst large enough for traditional machine learning algorithms (i.e. Support Vector Machines, Random Forest), does not meet the scale for a modern DNN, which number of parameters can easily exceed 10x the number of points available. The iQmulus is more suited in terms of size however due to a 2D semi-manual data labelling approach, many mislabelled ground truth points exist. 

A direct effort to address the need of large labelled datasets for deep learning algorithms is the Semantic3D dataset \cite{HackelEtAlT2017}. Semantic3D consists of $\sim$4BN points collected over 30 non-overlapping TLS scans. Points are classified into 8 categories; man made terrain, natural terrain, high vegetation, low vegetation, buildings, hardscape, scanning artefacts and cars. Although this dataset is certainly very valuable to the community, many caveats still exist. Most prominent is the use of a TLS over MLS. Scanning large areas with a static scanner is very time-consuming and ensuring complete coverage without significant redundancy is very difficult. As such, the Geomatics community has typically moved towards the use of MLS for large scale mapping, which is capable of scanning much larger areas in time scales orders of magnitudes less than static approaches. This causes issues when pre-training MLS data as TLS data exhibits very different artefacts to MLS. Furthermore, as with all the datasets discussed, strong class imbalances are present within the datasets. This is caused by the natural class imbalance of outdoor scenes. For example, a typical urban scan can consist of $>$90\% more road and fa{\c c}ade points over less prominent classes such as pole features and street furniture. This has been shown to negatively affect the network performance when training DNNs \cite{GriffithsBoehmD2019}. For a more thorough review on datasets available including indoor datasets we refer the reader to \cite{GriffithsBoehmD2019a}.

Although the ability to transfer learn from synthetic data is widely exploited for 2D images, this is less prominent in 3D point cloud learning. Wu et al., \cite{WuEtAlB2017, WuEtAlB2018b}, exploit the widely acclaimed Grand Theft Auto V game by creating a plugin to allow the mounting of a simulated Velodyne 64 scanner on top of a vehicle. Despite being able to train on a huge source of labelled synthetic data, their model SqueezeSeg generalised very poorly when switching to a real world domain, with a test accuracy of 29\%. This was accredited to \textit{dropout noise}, defined as missing points from the sensed point cloud caused by limited sensing range, mirror diffusion of the sensing laser, or jitter in the incident angles. SqueezeSegV2 proposed a domain adaption pipeline whereby dropout noise was mitigated by a Context Aggregation Module, increasing real world test accuracy to 57.4\%. This work is a strong influence for our dataset, as firstly the authors demonstrate the success of domain adaption techniques. Secondly, SqueezeSegV2 does not perform learning directly in the 3D space, but instead projects the points on to a spherical surface to allow for easier convolutions. The dataset is only available in the projected data format, and therefore can not be used for networks that operate directly on point clouds. 

The research field of autonomous driving also exploits synthetic data to create realistic agent training environments. This is largely beneficial as the vehicles can be developed in a safe environment where accidents costs nothing. The Synthia dataset \cite{RosEtAlG2016} is a vehicle drive through a virtual world. The dataset contains 2D imagery but also a lidar inspired 2.5D registered depth map. As the primary purpose of the research is to aid semantic segmentation for a moving vehicle, a full 3D point cloud is not released with the dataset. The authors do demonstrate the added benefit of pre-training on synthetic data. The more recent CARLA simulator \cite{DosovitskiyEtAlA2017} builds on Synthia, but is released as a full simulator software with a fully integrated API. Using CARLA, a globally registered 3D point cloud could be generated inside the very realistic and complex environment. 

\section{Generation}\label{generation}

The primary aim in generating our dataset is to produce a globally registered point cloud where each point $\mathbf{P}\in\mathbb{R}^{n\times3}$. Additionally, each point $\mathbf{P}$ contains a feature vector $\mathcal{F}\in \mathbb{R}^{n\times d}$ where $n$ is the number of points such that $n=367.9M$ and $d$ is red, green, blue, time, end of line, and label $l$ where $l \in L$ such that $|L|=9$.

\begin{figure}[t!]
	
	\includegraphics[width=0.5\textwidth]{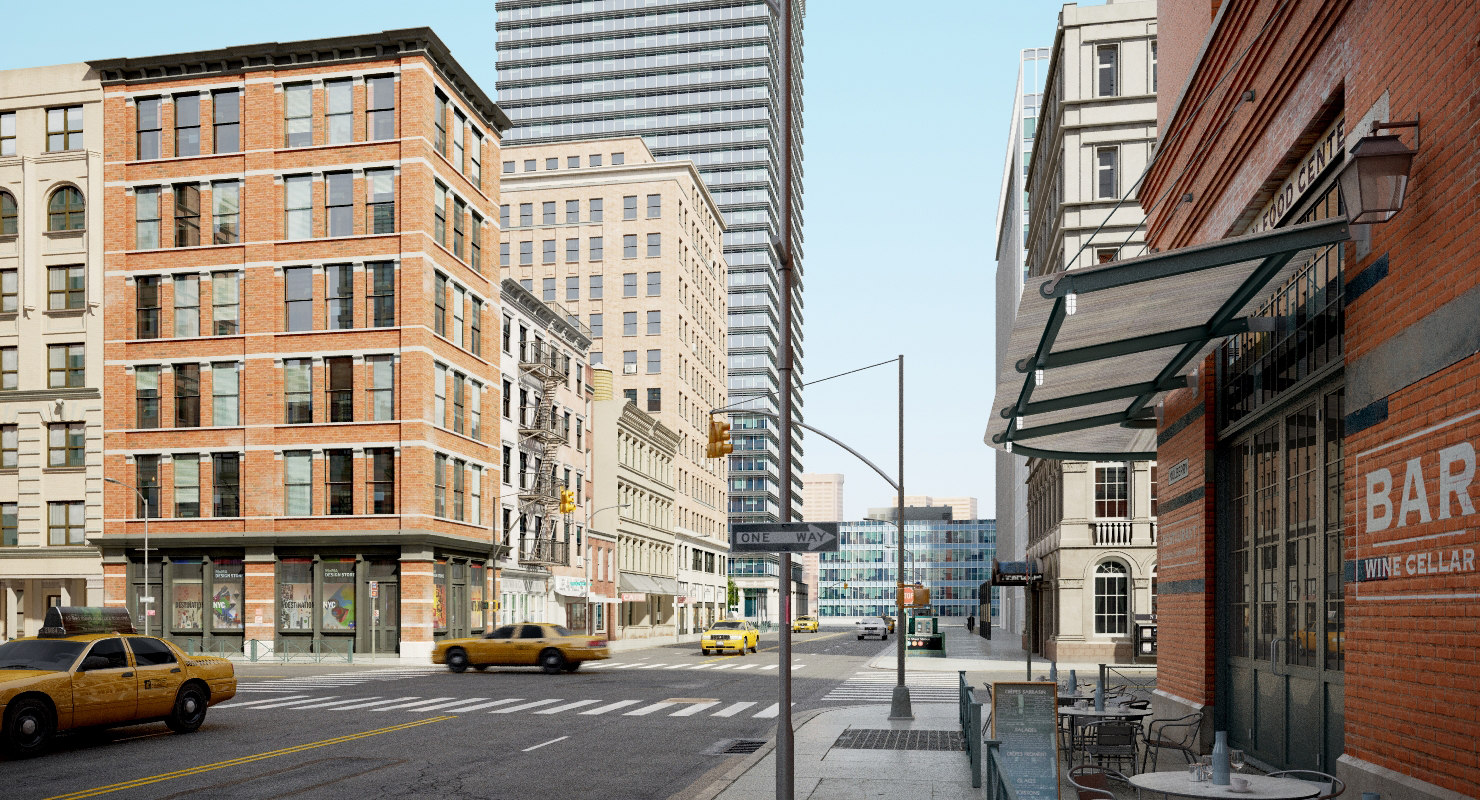}
	\caption{Rendered image from the initial downloaded model. Image source \cite{TurboSquid}.}
	\label{fig:stock}
	
\end{figure}

The SynthCity data was modelled inside the open-source Blender 3D graphics software \cite{BlenderOnlineCommunity2018}. The initial model was downloaded from an online model database (Fig. \ref{fig:stock}). The model was subsequently duplicated with the object undergoing shuffling to ensure the two areas were not identical to one another. Road segments were duplicated to connect the two urban environments leaving large areas of unoccupied space. To populate these areas additional typical suburban building models were downloaded and placed along the road. With respect to model numbers the dataset contains; 130 buildings, 196 cars, 21 natural ground planes, 12 ground planes, 272 pole-like objects, 172 road objects, 1095 street furniture objects and 217 trees (table \ref{tab:counts}). The total disk size of the model was 16.9GB. The primary restriction for the size of the dataset was availability of Random Access Memory (RAM) required on the workstation used for creating the model. This was limited to 32GB in our case, however, with a larger RAM the model size could have easily been extended.

The open-source Blender Sensor Simulation plugin Blensor \cite{GschwandtnerEtAlM2011} was used for simulation of the MLS and thus point cloud generation. We use the following setup for scanning: \\

\begin{tabular}{ll}
	
	Scan type & Generic lidar \\
	Max distance & 100m \\
	Angle resolution & 0.05m \\
	Start angle & \ang{-180} \\
	End angle & \ang{180} \\
	Frame time & $1/24s$ 
	
\end{tabular} \\

A typical scan took $\sim$330s to render and a total of 75,000 key frames were rendered from a pre-defined trajectory. To increase realism and generate more variability in point density the trajectory spline was moved by a random permutation at random intervals in all $x,y,z$ directions. The final rendering required $(330\times75000)/86400=286.46$ days CPU compute time. This was processed using AWS cloud computing service. We launched 22 type r4.2xlarge Ubuntu 18.04 EC2 spot instances, each containing 8 virtual CPUs and 61GB RAM. These were selected as rendering typically required $\sim$50GB RAM. All data was read and written to a EFS file storage system to allow for joint access of a single model instance. The total rendering time took $\sim$13 days on 22 EC2 instances.

Each render node produces an individual file $s_{t}$ for the 2D scan at time frame $t$. To create the global 3D point cloud each point must undergo a transformation $\mathbf{T}$ with respect to the scanner location $S_{x,y,z}$ and rotation $S_{\omega,\phi,\kappa}$. Blensor can export both $S_{x,y,z}$ and $S_{\omega,\phi,\kappa}$ at time $t$ as a motion file. Each scan is passed through a global registration script where the transformation $\mathbf{T}$ is computed as the rotation matrix where:

\begin{equation}
\mathbf{R_{x}}=
\begin{bmatrix}
1 & 0 & 0 \\
0 & \cos\omega & -sin\omega \\
0 & \sin\omega & \cos\omega
\end{bmatrix}
\end{equation}

\begin{equation}
\mathbf{R_{y}}=
\begin{bmatrix}
\cos\phi & 0 & \sin\phi \\
0 & 1 & 0 \\
-\sin\phi & 0 & \cos\phi
\end{bmatrix}
\end{equation}

\begin{equation}
\mathbf{R_{z}}=
\begin{bmatrix}
\cos\kappa & -sin\kappa & 0 \\
\sin\kappa & \cos\kappa & 0 \\
0 & 0 & 1
\end{bmatrix}
\end{equation}

\begin{equation}
\mathbf{R} = \mathbf{R_{z}}\mathbf{R_{y}}\mathbf{R_{x}}
\end{equation}

\begin{equation}
\mathbf{T} =
\begin{bmatrix}
R_{1,1} & R_{1,2} & R_{1,3} & S_{x}  \\
R_{2,1} & R_{2,2} & R_{2,3} & S_{y}  \\
R_{3,1} & R_{3,2} & R_{3,3} & S_{z}  \\
0 & 0 & 0 & 1
\end{bmatrix}
\end{equation}

Finally, each transformed point $\hat{p}_{t}$ is computed as:

\begin{equation}
\hat{p}_{t} = p_{t} \cdot \mathbf{T}
\end{equation}

In a separate post-processing stage we generate the features $\mathcal{F} = xn$,$yn,zn,time, eol$. To create $\mathcal{F} = xn$,$yn,zn$ we simply apply a 0.005m Gaussian noise to each $p_{x}$, $p_{y}$ and $p_{z}$ independently such that $pn_{i} = p_{i}^{x}+\sigma^{1},p_{i}^{y}+\sigma^{2},p_{i}^{z}+\sigma^{3}$ where $-0.005<\sigma<0.005$. We choose 0.005m as this is in line with an expected modern scanner noise at this distance. $\mathcal{F} = time$ is calculated by adding the key frame time available in the motion file with the scanner point time available in the individual scan files. This is effectively a simulated GNSS time available with MLS and ALS point clouds. Finally the end of line (eol) is calculated as a binary indicator where the $p_{i}^{eol}=1$ if it is the final point acquired by the individual scan $s_{t}$ or 0 otherwise.

\begin{table}[ht]
	\centering
	\begin{tabular}{ll}
		\textbf{Feature} & \textbf{Type} \\
		\hline
		$x$ & float \\
		$y$ & float \\
		$z$ & float \\
		$x_{n}$ & float \\
		$y_{n}$ & float \\
		$z_{n}$ & float \\
		R & short \\
		G & short \\
		B & short \\
		time & double \\
		eol & boolean [0, 1] \\
		label & short [0-8] \\
		\hline	
	\end{tabular}
	\caption{Data fields and stored types.}
	\label{tab:fields}
\end{table}

We choose to store our data in the parquet data format \cite{parquet}. The parquet format is very efficient with respect to memory storage but is also very suitable for out-of-memory processing. The parquet format is designed to integrate with the Apache Hadoop ecosystem. It can be directly read into python Pandas dataframes but also python Dask data frames which allow for easy out-of-memory processing directly in the python ecosystem. 

\section{Data}

The dataset is modelled from a completely fictional typical urban environment. In reality the environment would be most similar to that of downtown and suburban New York City, USA. This was due to the initial starting model, and not any design choices made by ourselves. Other buildings and street infrastructure are typical of mainland Europe. We classify each point into one category from; road, pavement, ground, natural ground, tree, building, pole-like, street furniture or car. To address the class imbalance issue, during construction of the model we aimed to bias the placement of small less dominant features in an attempt to reduce this as much as possible. As point cloud DNNs typically work on small subsets of the dataset we argue that this approach should not introduce any unfavourable bias, but instead help physically reduce the class imbalance.

\begin{table}[ht]
	\centering
	
	\begin{tabular}{lrr}
		\textbf{Label} & \textbf{No. Models} & \textbf{No. Points} \\
		\hline
		Road & 172 & 215,870,472 \\
		Pavement & 172 & 21,986,017 \\
		Ground & 12 & 6,206,312 \\
		Natural ground & 21 & 4,788,775 \\
		Tree & 217 & 12,088,077 \\
		Building & 130 & 97,973,820 \\
		Pole-like & 272 & 1,636,443 \\
		Street furniture & 1095 & 1,469,766 \\
		Car & 196 & 5,907,347 \\
		\hline
		Total & 2287 & 367,927,029 \\
		\hline
	\end{tabular}
	\caption{Label categories and number of points per category.}
	\label{tab:counts}
\end{table}

The final feature list with their respective storage type is shown in table \ref{tab:fields}. The total number of points generated is shown in table \ref{tab:counts}. The disk space of the complete parquet file is 27.5GB, as a typical work station would not be able to load this model into memory, we split this scan into 9 sub areas. Each sub area is split solely on horizontal coordinates and can therefore contain points from any scan at any key frame. The purpose of this is twofold; firstly, users of typical workstations can load an area directly into memory, and secondly, we can nominate a fixed test area. We propose that areas 1-2 and 4-9 be used for training and area 3 be reserved for model testing. This enables consistency if models trained on our dataset are to be compared from one another. We choose area 3 as it contains a good representation of all classes. As SynthCity is not designed as a benchmark dataset we provide the ground truth labels for area 3 in the same manner as all other areas.

\begin{figure}[h!]
	\includegraphics[width=0.5\textwidth]{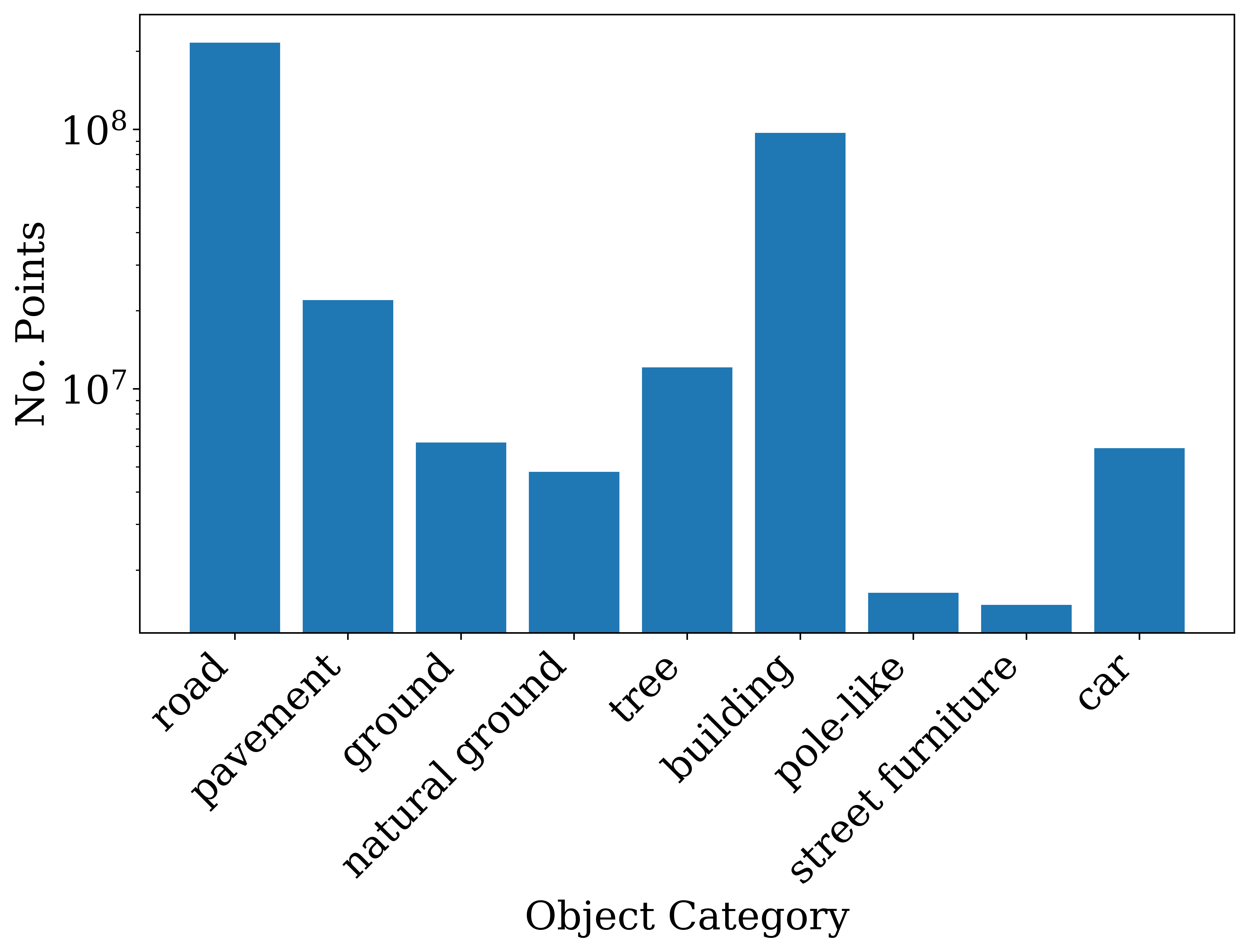}
	\caption{Total point counts for each label category. Note the log y-axis scale.}
	\label{fig:bar}
\end{figure}

\section{Discussion}

Although SynthCity was modelled to be biased toward poorly represented categories (i.e. street furniture and pole-like objects), it is evident that a significant class imbalance still exists (Fig. \ref{fig:bar}). The reasons for this is twofold. Firstly, continuous features such as road and pavement cover significantly larger areas than smaller discrete features. Secondly, due to the nature of MLS, objects closer to the scanner are sampled with a higher point density. As MLS are typically car mounted, road and pavement naturally have very high point densities. A sensible pre-processing approach to account for this issue is to first voxel downsample the point cloud to a regular point density. This technique has been shown to considerably improve classification accuracy for both outdoor and indoor point clouds \cite{GriffithsBoehmD2019}. As one of the primary benefits of a self-constructed synthetic model is the ability to choose the object placement distribution, it is evident from our dataset that this should be further exaggerated still. 

SynthCity has been designed primarily to be used for semantic per-point classification. As such each point contains a feature vector and a classification label. Whilst this is useful for a range of applications, currently the dataset does not contain instance id's for individual object extraction. As each object is a discrete object within the Blender environment extraction of instance id's would be reasonably trivial to extract. Moreover, a simple post processing script could be employed to convert instance id's to 3D instance bounding boxes which would enable the dataset to be used for 3D object localisation algorithms as well as per-point classification. With SynthCity being an ongoing project we plan to implement this in future releases.

Blensor supports the ability scan with a range of scanners, most notably a simulated Velodyne scanner. Such scanners are commonly used for both MLS systems and autonomous vehicles. Re-rendering with a Velodyne scanner would only require the AWS instances to be run again to produce the equivalent point cloud. Furthermore, scanner properties can be changed to simulate a range of scanners that are currently not covered by the pre-defined settings. We argue that as with 2D images, 3D point clouds should be sensor invariant. Training on multiple sensors would likely be a very valuable augmentation technique.

\section{Conclusion}

In this work we present SynthCity an open, large-scale synthetic point cloud. We release this dataset to help aid research in the potential use for pre-training of segmentation/classification models on synthetic datasets. We argue an ability to generalise from synthetic data to real world data would be immensely beneficial to the community as such a wealth of existing synthetic 3D environments already exist. Most notably those generated from the gaming, virtual environment and simulated training industries. Our model contains 367.9M perfect labelled points with 5 additional features; red, green, blue, time, eol. In addition we also present an identical point cloud with the permutation of Gaussian sampled noise, giving the point cloud a more realistic appearance.

\section*{Acknowledgements}

This work incorporates results from the research project "Open ML Training Data For Visual Tagging Of Construction-specific Objects (ConTag)" funded by the Centre for Digital Built Britain (CDBB), under InnovateUK grant number RG96233. 

\bibliographystyle{IEEEtran}
\bibliography{references}

% Generated by IEEEtran.bst, version: 1.14 (2015/08/26)
\begin{thebibliography}{10}
\providecommand{\url}[1]{#1}
\csname url@samestyle\endcsname
\providecommand{\newblock}{\relax}
\providecommand{\bibinfo}[2]{#2}
\providecommand{\BIBentrySTDinterwordspacing}{\spaceskip=0pt\relax}
\providecommand{\BIBentryALTinterwordstretchfactor}{4}
\providecommand{\BIBentryALTinterwordspacing}{\spaceskip=\fontdimen2\font plus
\BIBentryALTinterwordstretchfactor\fontdimen3\font minus
  \fontdimen4\font\relax}
\providecommand{\BIBforeignlanguage}[2]{{%
\expandafter\ifx\csname l@#1\endcsname\relax
\typeout{** WARNING: IEEEtran.bst: No hyphenation pattern has been}%
\typeout{** loaded for the language `#1'. Using the pattern for}%
\typeout{** the default language instead.}%
\else
\language=\csname l@#1\endcsname
\fi
#2}}
\providecommand{\BIBdecl}{\relax}
\BIBdecl

\bibitem{DengJ2009}
J.~Deng, W.~Dong, R.~Socher, L.-J. Li, K.~Li, and L.~{Fei-Fei}, ``Imagenet:
  {{A}} large-scale hierarchical image database,'' in \emph{Computer {{Vision}}
  and {{Pattern Recognition}}, 2009. {{CVPR}} 2009. {{IEEE Conference}}
  On}.\hskip 1em plus 0.5em minus 0.4em\relax IEEE, 2009, pp. 248--255.

\bibitem{LinEtAlT2014}
T.-Y. Lin, M.~Maire, S.~Belongie, J.~Hays, P.~Perona, D.~Ramanan,
  P.~Doll{\'a}r, and C.~L. Zitnick, ``\BIBforeignlanguage{en}{Microsoft
  {{COCO}}: {{Common Objects}} in {{Context}}},'' in
  \emph{\BIBforeignlanguage{en}{Computer {{Vision}} \textendash{} {{ECCV}}
  2014}}, ser. Lecture {{Notes}} in {{Computer Science}}, D.~Fleet, T.~Pajdla,
  B.~Schiele, and T.~Tuytelaars, Eds.\hskip 1em plus 0.5em minus 0.4em\relax
  {Springer International Publishing}, 2014, pp. 740--755.

\bibitem{EveringhamEtAlM2010}
M.~Everingham, L.~Van~Gool, C.~K.~I. Williams, J.~Winn, and A.~Zisserman,
  ``\BIBforeignlanguage{en}{The {{Pascal Visual Object Classes}} ({{VOC}})
  {{Challenge}}},'' \emph{\BIBforeignlanguage{en}{International Journal of
  Computer Vision}}, vol.~88, no.~2, pp. 303--338, Jun. 2010.

\bibitem{ChangEtAlA2015b}
A.~X. Chang, T.~Funkhouser, L.~Guibas, P.~Hanrahan, Q.~Huang, Z.~Li,
  S.~Savarese, M.~Savva, S.~Song, H.~Su, J.~Xiao, L.~Yi, and F.~Yu,
  ``{{ShapeNet}}: {{An Information}}-{{Rich 3D Model Repository}},''
  \emph{arXiv:1512.03012 [cs]}, Dec. 2015.

\bibitem{SernaEtAlA2014}
A.~Serna, B.~Marcotegui, F.~Goulette, and J.-E. Deschaud,
  ``Paris-rue-{{Madame}} database: A {{3D}} mobile laser scanner dataset for
  benchmarking urban detection, segmentation and classification methods,'' in
  \emph{4th {{International Conference}} on {{Pattern Recognition}},
  {{Applications}} and {{Methods ICPRAM}} 2014}, {Angers, France}, Mar. 2014.

\bibitem{ValletEtAlB2015}
B.~Vallet, M.~Br{\'e}dif, A.~Serna, B.~Marcotegui, and N.~Paparoditis,
  ``{{TerraMobilita}}/{{iQmulus}} urban point cloud analysis benchmark,''
  \emph{Computers \& Graphics}, vol.~49, pp. 126--133, Jun. 2015.

\bibitem{RoynardEtAlX2017}
X.~Roynard, J.-E. Deschaud, and F.~Goulette, ``Paris-{{Lille}}-{{3D}}: A large
  and high-quality ground truth urban point cloud dataset for automatic
  segmentation and classification,'' \emph{arXiv:1712.00032 [cs, stat]}, Nov.
  2017.

\bibitem{HackelEtAlT2017}
T.~Hackel, N.~Savinov, L.~Ladicky, J.~D. Wegner, K.~Schindler, and
  M.~Pollefeys, ``{{Semantic3D}}.net: {{A}} new {{Large}}-scale {{Point Cloud
  Classification Benchmark}},'' \emph{arXiv:1704.03847 [cs]}, Apr. 2017.

\bibitem{GriffithsBoehmD2019}
D.~Griffiths and J.~Boehm, ``Weighted {{Point Cloud Augmentation}} for {{Neural
  Network Training Data Class}}-{{Imbalance}},'' \emph{ISPRS - International
  Archives of the Photogrammetry, Remote Sensing and Spatial Information
  Sciences}, vol. XLII-2/W13, pp. 981--987, Jun. 2019.

\bibitem{GriffithsBoehmD2019a}
------, ``\BIBforeignlanguage{en}{A {{Review}} on {{Deep Learning Techniques}}
  for {{3D Sensed Data Classification}}},''
  \emph{\BIBforeignlanguage{en}{Remote Sensing}}, vol.~11, no.~12, p. 1499,
  Jan. 2019.

\bibitem{WuEtAlB2017}
B.~Wu, A.~Wan, X.~Yue, and K.~Keutzer, ``{{SqueezeSeg}}: {{Convolutional Neural
  Nets}} with {{Recurrent CRF}} for {{Real}}-{{Time Road}}-{{Object
  Segmentation}} from {{3D LiDAR Point Cloud}},'' \emph{arXiv:1710.07368 [cs]},
  Oct. 2017.

\bibitem{WuEtAlB2018b}
B.~Wu, X.~Zhou, S.~Zhao, X.~Yue, and K.~Keutzer, ``{{SqueezeSegV2}}: {{Improved
  Model Structure}} and {{Unsupervised Domain Adaptation}} for
  {{Road}}-{{Object Segmentation}} from a {{LiDAR Point Cloud}},''
  \emph{arXiv:1809.08495 [cs]}, Sep. 2018.

\bibitem{RosEtAlG2016}
G.~Ros, L.~Sellart, J.~Materzynska, D.~Vazquez, and A.~M. Lopez, ``The
  {{SYNTHIA Dataset}}: {{A Large Collection}} of {{Synthetic Images}} for
  {{Semantic Segmentation}} of {{Urban Scenes}},'' in \emph{Proceedings of the
  {{IEEE Conference}} on {{Computer Vision}} and {{Pattern Recognition}}},
  2016, pp. 3234--3243.

\bibitem{DosovitskiyEtAlA2017}
A.~Dosovitskiy, G.~Ros, F.~Codevilla, A.~Lopez, and V.~Koltun, ``{{CARLA}}:
  {{An Open Urban Driving Simulator}},'' \emph{arXiv:1711.03938 [cs]}, Nov.
  2017.

\bibitem{BlenderOnlineCommunity2018}
B.~O. Community, ``Blender - a {{3D}} modelling and rendering package,''
  Blender Foundation, 2018.

\bibitem{TurboSquid}
T.~Squid, ``Metropolis {{City Experience}},''
  https://www.turbosquid.com/3d-models/city-modular-new-max/982288, accessed:
  2019-07-10.

\bibitem{GschwandtnerEtAlM2011}
M.~Gschwandtner, R.~Kwitt, A.~Uhl, and W.~Pree,
  ``\BIBforeignlanguage{en}{{{BlenSor}}: {{Blender Sensor Simulation
  Toolbox}}},'' in \emph{\BIBforeignlanguage{en}{Advances in {{Visual
  Computing}}}}, ser. Lecture {{Notes}} in {{Computer Science}}, G.~Bebis,
  R.~Boyle, B.~Parvin, D.~Koracin, S.~Wang, K.~Kyungnam, B.~Benes, K.~Moreland,
  C.~Borst, S.~DiVerdi, C.~{Yi-Jen}, and J.~Ming, Eds.\hskip 1em plus 0.5em
  minus 0.4em\relax {Springer Berlin Heidelberg}, 2011, pp. 199--208.

\bibitem{parquet}
``Apache {{Parquet}},'' https://parquet.apache.org/, accessed: 2019-07-10.

\end{thebibliography}

\end{document}